\documentclass{bmvc2k}

\newif\ifanonymous
\anonymousfalse
\ifanonymous
\bmvcreviewcopy{383}
\fi

\title{Beyond String Matching: Semantic Evaluation of PDF Table Extraction}

\addauthor{Pius Horn}{pius.horn@hs-offenburg.de}{1}
\addauthor{Janis Keuper}{}{1,2}

\addinstitution{
 Institute for Machine Learning and Analytics (IMLA)\\
 Offenburg University\\
 Offenburg, Germany
}
\addinstitution{
 University of Mannheim\\
 Mannheim, Germany
}

\runninghead{Horn, Keuper}{Benchmarking Document Parsers on Table Extraction}

\usepackage{amsfonts}
\usepackage{amssymb}
\usepackage{float}
\usepackage{booktabs}
\usepackage{xcolor}
\usepackage{colortbl}
\usepackage{multirow}
\usepackage{hhline}
\usepackage{makecell}
\usepackage{tabularray}
\usepackage{tikz}

\definecolor{cheader}{HTML}{E2E8F0}      % light gray  - table headers
\definecolor{clayout}{HTML}{FEF3C3}      % warm yellow - layout (header-flattening, rowspan)
\definecolor{cnotation}{HTML}{BFDBFE}    % clear blue  - representational equivalence (symbol- and value-level)
\definecolor{cartifact}{HTML}{D5E8D4}    % soft green  - markup artifact (bold encoding)
\definecolor{ccontent}{HTML}{FECACA}     % subtle red  - content errors

\definecolor{tabblue}{RGB}{31,119,180}
\definecolor{tabred}{RGB}{214,39,40}

\newlength{\cellw}
\newlength{\barw}
\newcommand{\sbar}[3]{\textcolor{#3}{\rule{#1\barw}{1mm}}\textcolor{black!15}{\rule{#2\barw}{1mm}}}

\newcommand{\hl}[2]{{\setlength{\fboxsep}{1pt}\colorbox{#1}{#2}}}
\begin{document}

\maketitle

\begin{abstract}
Reliably extracting tables from PDFs is essential for large-scale scientific data mining and knowledge base construction, yet existing evaluation approaches rely on rule-based metrics that fail to capture semantic equivalence of table content.
We present a benchmarking framework based on synthetically generated PDFs with precise LaTeX ground truth, using tables sourced from arXiv to ensure realistic complexity and diversity.
As our central methodological contribution, we apply LLM-as-a-judge for semantic table evaluation, integrated into a matching pipeline that accommodates inconsistencies in parser outputs.
Through a human validation study comprising over 1,500 quality judgments on extracted table pairs, we show that LLM-based evaluation achieves substantially higher correlation with human judgment (Pearson r=0.93) compared to currently used Tree Edit Distance-based Similarity (TEDS, r=0.68) and Grid Table Similarity (GriTS, r=0.70).
Evaluating 21 contemporary PDF parsers across 100 synthetic documents containing 451 tables reveals significant performance disparities.
Our results offer practical guidance for selecting parsers for tabular data extraction and establish a reproducible, scalable evaluation methodology for this critical task.
\end{abstract}

%-------------------------------------------------------------------------
\section{Introduction}
\label{sec:intro}

\begin{figure}[!t]
\centering
% =====================================================================
% Three panels in a single row:
% (a) problem  (b) solution  (c) dataset + parser scores
% =====================================================================
\begin{minipage}[t]{0.32\textwidth}
\centering
\footnotesize\textbf{(a) Rule-based metrics\\ mislead}\\[1mm]
\begin{minipage}[c][40mm][c]{\linewidth}
\centering
{\scriptsize
\setlength{\fboxsep}{1.5pt}
\setlength{\aboverulesep}{2pt}
\setlength{\belowrulesep}{2pt}
\renewcommand{\arraystretch}{1.2}
\settowidth{\cellw}{\texttt{\textbackslash alpha}}
\setlength{\barw}{\dimexpr\cellw + 2\fboxsep + 2\fboxrule\relax}
\begin{tabular}{@{}l@{\hspace{1mm}}c@{\hspace{1mm}}c@{}}
       & \textit{semantic}   & \textit{benign}     \\[-1.1mm]
       & \textit{errors}     & \textit{variations} \\[0.4mm]
GT     & \fcolorbox{black!40}{ccontent!60}{\strut\makebox[\cellw][c]{\texttt{1.12}}} & \fcolorbox{black!40}{cnotation!60}{\strut\makebox[\cellw][c]{\texttt{\textbackslash alpha}}} \\[-0.4mm]
       & $\downarrow$ & $\downarrow$ \\[-0.4mm]
Parsed & \fcolorbox{black!40}{ccontent!60}{\strut\makebox[\cellw][c]{\texttt{112}}} & \fcolorbox{black!40}{cnotation!60}{\strut\makebox[\cellw][c]{$\alpha$}} \\
\midrule
\makecell[l]{\textit{Score}\\\textit{should be}}\rule{0pt}{2.4ex} & \textcolor{red!70!black}{\textbf{low}} & \textcolor{green!55!black}{\textbf{high}} \\
\makecell[l]{\textbf{Human}\\\textbf{Reference}}\rule{0pt}{2.4ex} & \sbar{0.20}{0.80}{red!70!black} & \sbar{0.95}{0.05}{green!55!black} \\
\makecell[l]{Rule-based\\{\scriptsize TEDS/GriTS}}\rule{0pt}{2.4ex} & \sbar{0.8}{0.2}{green!55!black} & \sbar{0.21}{0.79}{red!70!black} \\
\end{tabular}}
\end{minipage}\\[0.5mm]
{\scriptsize
\begin{tabular}{@{}c@{}}
rule-based score $\not\Rightarrow$\\[-0.2mm]
extraction quality
\end{tabular}}
\end{minipage}\hfill
\begin{minipage}[t]{0.32\textwidth}
\centering
\footnotesize\textbf{(b) LLM judge\\ aligns with humans}\\[1mm]
\begin{minipage}[c][40mm][c]{\linewidth}
\centering
\begin{tikzpicture}[x=3.15mm,y=2.75mm]
% Axes
\draw[->, black!60, line width=0.3pt] (0,0) -- (10.6,0);
\draw[->, black!60, line width=0.3pt] (0,0) -- (0,10.6);
\foreach \v in {0,5,10} {
  \draw[black!60, line width=0.3pt] (\v,-0.15) -- (\v,0.15);
  \draw[black!60, line width=0.3pt] (-0.15,\v) -- (0.15,\v);
  \node[font=\scriptsize, black!70, anchor=north] at (\v,-0.2) {\v};
  \node[font=\scriptsize, black!70, anchor=east] at (-0.25,\v) {\v};
}
\node[font=\scriptsize, black!70, anchor=north] at (5,-1.2) {human score};
\node[font=\scriptsize, black!70, anchor=south, rotate=90] at (-0.9,5) {metric score};
% Diagonal reference
\draw[gray!50, dashed, line width=0.3pt] (0,0) -- (10,10);
% Clip clouds/lines to plot area
\begin{scope}
\clip (-0.2,-0.2) rectangle (10.3,10.3);
% TEDS (rule-based, blue) — std cloud + mean line + bin dots
\fill[tabblue, opacity=0.18] (0,2.66) -- (1,5.79) -- (2,7.65) -- (3,8.86) -- (4,8.76) -- (5,9.33) -- (6,9.31) -- (7,9.54) -- (8,9.20) -- (9,9.56) -- (10,10.00) -- (10,8.63) -- (9,6.86) -- (8,6.86) -- (7,5.86) -- (6,6.72) -- (5,5.43) -- (4,6.17) -- (3,6.23) -- (2,4.19) -- (1,2.14) -- (0,0.40) -- cycle;
\draw[tabblue, line width=0.6pt, line join=round]
  (0,1.53) -- (1,3.97) -- (2,5.92) -- (3,7.55) -- (4,7.46) -- (5,7.38) -- (6,8.02) -- (7,7.70) -- (8,8.03) -- (9,8.21) -- (10,9.46);
\foreach \x/\y in {0/1.53, 1/3.97, 2/5.92, 3/7.55, 4/7.46, 5/7.38, 6/8.02, 7/7.70, 8/8.03, 9/8.21, 10/9.46} {
  \fill[tabblue] (\x,\y) circle (0.4mm);
}
% Gemma (LLM judge, red) — std cloud + mean line + bin dots
\fill[tabred, opacity=0.18] (0,1.21) -- (1,2.08) -- (2,2.61) -- (3,4.33) -- (4,4.89) -- (5,5.45) -- (6,6.99) -- (7,9.16) -- (8,9.55) -- (9,10.00) -- (10,10.00) -- (10,9.49) -- (9,7.48) -- (8,6.22) -- (7,5.47) -- (6,3.43) -- (5,1.77) -- (4,1.69) -- (3,1.30) -- (2,1.23) -- (1,0.49) -- (0,0.46) -- cycle;
\draw[tabred, line width=0.6pt, line join=round]
  (0,0.83) -- (1,1.29) -- (2,1.92) -- (3,2.81) -- (4,3.29) -- (5,3.61) -- (6,5.21) -- (7,7.31) -- (8,7.88) -- (9,8.80) -- (10,9.90);
\foreach \x/\y in {0/0.83, 1/1.29, 2/1.92, 3/2.81, 4/3.29, 5/3.61, 6/5.21, 7/7.31, 8/7.88, 9/8.80, 10/9.90} {
  \fill[tabred] (\x,\y) circle (0.4mm);
}
\end{scope}
% Inline labels — rule-based (blue) upper-left near its high line, LLM judge (red) lower-right near its diagonal line
\node[font=\tiny, tabblue, anchor=north west, align=left] at (0.3,11.3) {\textbf{rule-based}\\TEDS, $r{=}0.68$};
\node[font=\tiny, tabred, anchor=south east, align=right] at (10.3,0.3) {\textbf{LLM judge}\\Gemma-4-31B, $r{=}0.93$};
\end{tikzpicture}
\end{minipage}\\[0.5mm]
{\scriptsize validated on 518 pairs\\[-0.2mm]with 1{,}554 human ratings}
\end{minipage}\hfill
\begin{minipage}[t]{0.32\textwidth}
\centering
\footnotesize\textbf{(c) Synthetic benchmark\\ with real tables}\\[1mm]
\begin{minipage}[c][40mm][c]{\linewidth}
\centering
% PDF stack: two offset rectangles behind the front page give the
% illusion of a stack of documents. The \fill[white] before the
% \includegraphics is required because data/sample_page.pdf has a transparent
% background — without it, the back pages would shine through.
\begin{tikzpicture}[x=1mm,y=1mm]
\draw[fill=white, draw=black!25, line width=0.2pt] (3.5,3.5) rectangle (27.5,37.5);
\draw[fill=white, draw=black!35, line width=0.2pt] (1.75,1.75) rectangle (25.75,35.75);
\fill[white] (0,0) rectangle (24,34);
\node[anchor=south west, inner sep=0pt] at (0,0)
    {\includegraphics[width=24mm]{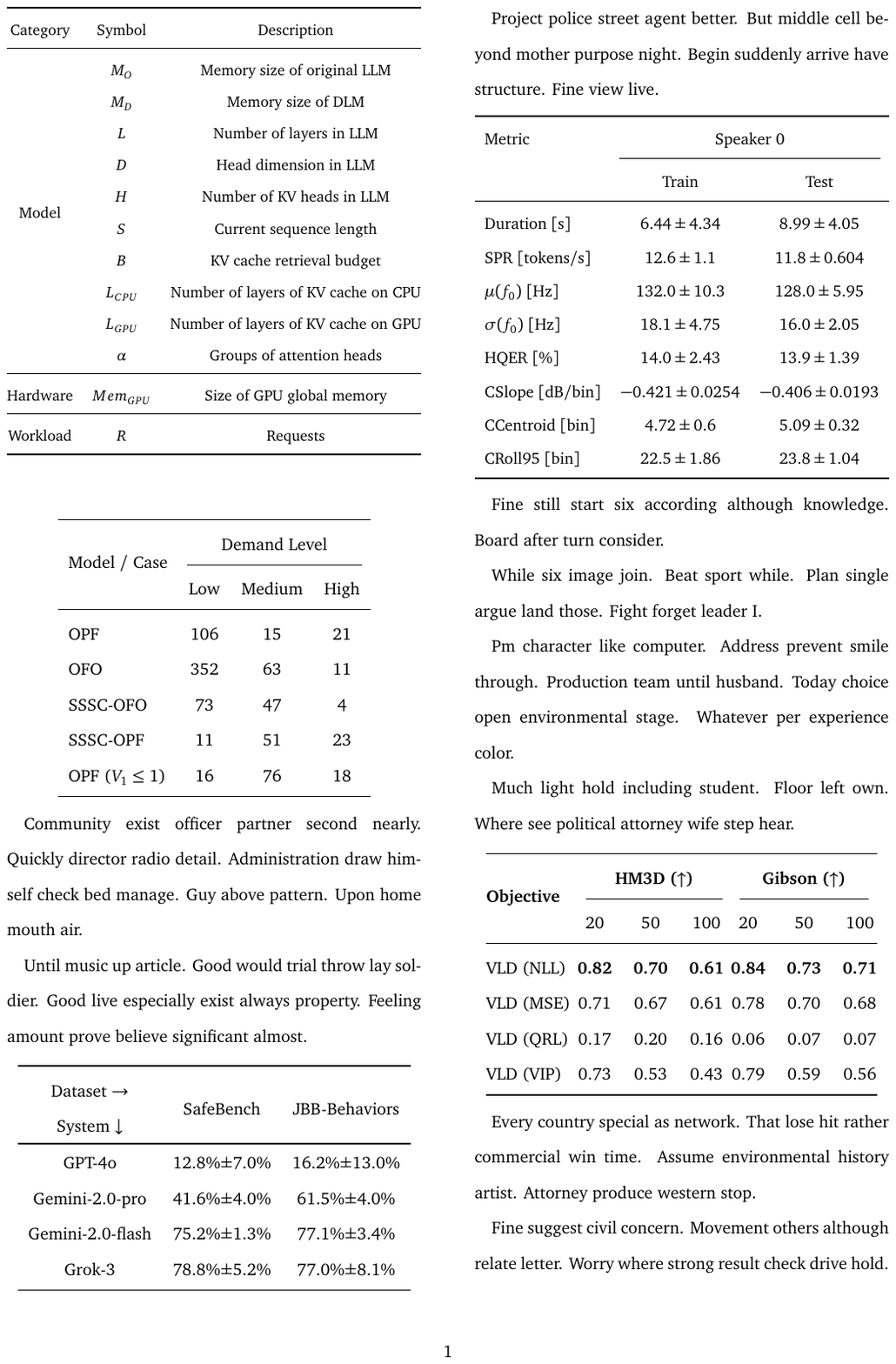}};
\draw[draw=black!80, line width=0.25pt] (0,0) rectangle (24,34);
\end{tikzpicture}
\end{minipage}\\[0.5mm]
{\scriptsize regeneratable,\\[-0.2mm]no manual annotation}
\end{minipage}
\vspace{5mm}
\caption{Rule-based metrics both over-reward semantically broken outputs and penalize benign representational variation (a); an LLM-as-a-judge approach validated on 1{,}554 human ratings aligns substantially better with human judgment (Pearson $r{=}0.93$ vs.\ $r{=}0.68$ for TEDS) (b); and used to benchmark 21 parsers on 100 synthetic PDFs that embed 451 real arXiv tables with their LaTeX source as exact ground truth (c).}
\label{fig:teaser_overview}
\end{figure}

Much of the structured knowledge in scientific publications, financial reports, and technical documents is organized in tables.
As document parsing becomes central to language model pretraining, retrieval-augmented generation, and scientific data mining~\cite{zhang2024parsing,olmocr2025}, the ability to accurately and reliably extract tabular data from PDFs has become increasingly important.

The landscape of PDF document parsing has evolved rapidly, with approaches ranging from rule-based extraction tools and specialized OCR models to end-to-end vision-language models~\cite{salaheldin2024deep,adhikari2024comparative}.
Existing benchmarks evaluate table extraction at scales from cropped table images~\cite{pubtabnet,pubtables1m} to full document-level assessments~\cite{omnidocbench,OmniOCRBenchmark}, and the accompanying metrics have progressed from cell adjacency relations~\cite{gobel2013icdar} to tree-based~\cite{pubtabnet} and grid-based~\cite{grits} comparison (see Section~\ref{sec:metrics}).
Yet all of these approaches rely on structural matching and surface-level string comparison, unable to assess whether the actual information conveyed by a table has been correctly preserved.
Consequently, a parser that produces a structurally different but semantically equivalent representation may be penalized unfairly, while one that preserves structure but corrupts cell content may receive an inflated score.

The LLM-as-a-judge paradigm~\cite{zheng2023judging} offers a promising solution, having demonstrated effectiveness for evaluating complex outputs where traditional metrics fall short~\cite{llm_judge}.
For table assessment, where content correctness and structural fidelity must be jointly evaluated, LLM-based evaluation can capture semantic nuances that surface-level similarity metrics miss.
% Camera-ready: this example is dropped to keep the introduction on pages 1--2,
\ifanonymous For instance, the Greek letter rendered as \texttt{\textbackslash alpha} versus the Unicode glyph $\alpha$, or a multi-level header flattened into concatenated single-row cells, is semantically identical to its ground truth but maximally different at the character level---exactly the kind of distinction an LLM judge can resolve while a string metric cannot.\fi

We combine this evaluation approach with a benchmarking framework that embeds real arXiv tables into synthetic PDFs, providing exact LaTeX ground truth without manual annotation.
Together, these contributions address both the metric and benchmark gaps:
\begin{itemize}
    \item \textbf{LLM-as-a-judge for semantic table evaluation}, reaching Pearson $r{=}0.93$ with human judgment versus $r{\leq}0.70$ for the rule-based metrics TEDS, GriTS, and SCORE, and usable as a drop-in metric on existing table-extraction benchmarks.
    \item \textbf{A meta-evaluation resource} of 1,554 human ratings over 518 table pairs, used to validate our metric and enabling assessment of future table extraction metrics.
    \item \textbf{A synthetic PDF benchmark} embedding real arXiv tables that post-date all evaluated parsers, providing exact LaTeX ground truth without manual annotation, paired with an LLM-based matching pipeline that aligns parser outputs to ground truth across diverse output formats.
    \item \textbf{A public leaderboard of 21 contemporary parsers}, spanning specialized OCR models, general-purpose VLMs, and rule-based tools, on 100 pages containing 451 tables, revealing substantial performance disparities and offering practical guidance for practitioners.
\end{itemize}

\section{Related Work}
\subsection{Document Parsing Benchmarks}
Existing benchmarks for table extraction fall into two broad categories: \emph{datasets of cropped table images} for isolated table recognition, and \emph{datasets of full document pages} where tables appear alongside text and figures.

\textbf{Datasets of cropped table images} have driven progress in table structure recognition from isolated images.
Large-scale resources include PubTabNet~\cite{pubtabnet} and PubTables-1M~\cite{pubtables1m}, both derived from PubMed Central scientific articles, FinTabNet~\cite{fintabnet} from financial reports, TableBank~\cite{tablebank} via weak supervision from Word and LaTeX documents, and SynthTabNet~\cite{synthtabnet} as a synthetically generated dataset, while SciTSR~\cite{scitsr} derives structure labels from LaTeX sources.
These datasets also shaped the dominant evaluation methodology: PubTabNet introduced TEDS, and GriTS~\cite{grits} later proposed grid-level evaluation as an alternative.
However, cropped-table datasets remove important page-level aspects of document parsing, including table localization within surrounding content and extraction from complete page layouts.
They are therefore well suited for table recognition in controlled settings, but only partially capture the end-to-end document parsing problem.

\textbf{Datasets of full document pages} evaluate table extraction from complete pages where tables appear alongside text and figures.
The OmniAI OCR Benchmark~\cite{OmniOCRBenchmark} evaluates overall document extraction accuracy but lacks table-specific metrics, while olmOCR-Bench~\cite{olmocr2025} and ParseBench~\cite{zhang2026parsebench} adopt a unit-test paradigm with binary pass/fail rules at the cell level rather than holistic table quality assessment.
OmniDocBench~\cite{omnidocbench} and READoc~\cite{readoc2025} (1,355 and 3,576~pages, respectively) go further by including explicit table evaluation, yet both rely on TEDS and edit distance metrics that capture structural similarity without assessing semantic equivalence.
PubTables-v2~\cite{pubtables_v2} provides a large-scale benchmark (467K~pages with 548K~tables) that extends PubTables-1M from cropped tables to full-page extraction, evaluated using GriTS.
Soric et al.~\cite{soric2025benchmarking} (approximately 84K~pages) propose new end-to-end extraction metrics alongside TEDS and GriTS, though all remain at the structural and string level.

\subsection{Table Extraction Evaluation Metrics}
\label{sec:metrics}
Evaluating extracted tables against ground truth requires metrics that jointly assess structural fidelity and content correctness.
Directed adjacency relations (DAR)~\cite{gobel2013icdar}, introduced in the ICDAR 2013 Table Competition as the first metric designed specifically for table structure evaluation, captured local cell neighborhoods but could not represent global table structure, motivating the tree-level and grid-level approaches that followed.

\textbf{Tree Edit Distance-based Similarity (TEDS).}
TEDS~\cite{pubtabnet}, introduced alongside PubTabNet, has become the de facto standard for table recognition evaluation.
Both predicted and ground-truth tables are represented as HTML trees whose leaf nodes (\texttt{<td>}) carry colspan, rowspan, and character-level tokenized content.
The tree edit distance is computed with a unit cost for structural mismatches and normalized Levenshtein distance for cell content, yielding $\mathrm{TEDS} = 1 - d\,/\,\max(|T_\mathrm{pred}|, |T_\mathrm{gt}|)$, where $d$ is the edit distance and $|T|$ the number of nodes in each tree.
Since both structure and content are compared at the character level, the score is sensitive to markup choices (e.g., \texttt{<thead>} vs.\ \texttt{<tbody>}, or \texttt{<th>} vs.\ \texttt{<td>}) and surface-level string differences, so two valid HTML serializations of the same table can receive markedly different scores even when they encode identical information.

\textbf{Grid Table Similarity (GriTS).}
GriTS~\cite{grits} addresses the HTML sensitivity of TEDS by operating directly on the table's 2D grid representation, using factored row and column alignment via dynamic programming.
It defines separate metrics for structural topology ($\mathrm{GriTS}_\mathrm{Top}$, via intersection-over-union on relative span grids) and content ($\mathrm{GriTS}_\mathrm{Con}$, via longest common subsequence similarity), each yielding precision, recall, and F-score.
By avoiding the tree representation, GriTS treats rows and columns symmetrically and is robust to markup variations, though content comparison remains string-based.

\textbf{SCORE.}
Although SCORE~\cite{score2025} presents itself as a ``semantic evaluation framework'' addressing the format rigidity of TEDS and GriTS, it normalizes tables into format-agnostic cell tuples and evaluates \emph{index accuracy} by checking whether cells occupy correct grid positions and \emph{content accuracy} via edit distance on cell text, with tolerance for small row/column offsets.
This avoids penalizing markup differences across output formats, though the structural tolerance may also mask genuine errors, and the underlying cell comparison remains string-based, leaving true semantic equivalence (such as notational variants or equivalent value formats) unaddressed.

\textbf{Text-based metrics} such as Levenshtein edit distance~\cite{levenshtein} or BLEU~\cite{bleu} operate at a strictly lower level of granularity: while the metrics above preserve cell-level structure, text-based approaches flatten the table into a token sequence, making scores dependent on serialization order and unable to distinguish structural from content errors.

While LLM-based evaluation has recently shown promise for formula extraction from PDFs~\cite{horn2025formula}, substantially outperforming text-based, tree-based, and image-based metrics in correlation with human judgment, no comparable study exists for table extraction.
Tables pose distinct challenges beyond linear formula expressions: cell values must be unambiguously associated with their row and column headers, multi-level headers and merged cells encode hierarchical relationships, and equivalent information can be rendered in markedly different layouts (transposed orientations, flattened headers, alternative column groupings).
Whether semantic LLM-based assessment transfers to this richer structural setting therefore remains an open question, with no prior meta-evaluation against human reference scores to settle it.

\subsection{Remaining Gaps}
\label{sec:gaps}

Prior work leaves two gaps.
On the benchmark side, document-level benchmarks leave three open concerns.
\textbf{Data contamination}: PubMed-derived corpora such as PubTables-v2~\cite{pubtables_v2} overlap with standard OCR/VLM pretraining data.
\textbf{Ground-truth fidelity}: auto-derived ground truth suffers from conversion artifacts (e.g., READoc~\cite{readoc2025}).
\textbf{Table density}: OmniDocBench~\cite{omnidocbench}, olmOCR-Bench~\cite{olmocr2025}, the OmniAI OCR Benchmark~\cite{OmniOCRBenchmark}, and Soric et al.~\cite{soric2025benchmarking} are dominated by table-free pages.

On the metric side, TEDS, GriTS, SCORE, and text-based variants all operate at the structural or string level, leaving semantic equivalence of table content unassessed.
As Section~\ref{sec:metrics_limitations} shows, these metrics over-penalize representational variation while under-penalizing genuine content errors.

Our work addresses both.
On the benchmark side, we use synthetic PDFs whose LaTeX source serves as exact ground truth, embedding real arXiv tables to ensure realistic diversity.
On the metric side, we apply LLM-as-a-judge for semantic table evaluation.
Together with a broad comparison of 21~contemporary parsers, this yields a reproducible evaluation framework that we validate against human judgment in Section~\ref{sec:assessment}.

\section{Benchmark Setup}
\label{sec:methodology}

Our benchmark rests on two key design decisions: (1)~using real tables extracted from arXiv to ensure realistic diversity, and (2)~embedding them into synthetically generated PDFs to obtain exact ground truth without manual annotation.
This section describes the resulting benchmark construction and the matching pipeline that aligns parser outputs to ground truth tables.

\subsection{Synthetic PDFs with Ground Truth}
\label{sec:arxiv_extraction}

\begin{figure}[!hb]
\centering
\includegraphics[width=\textwidth]{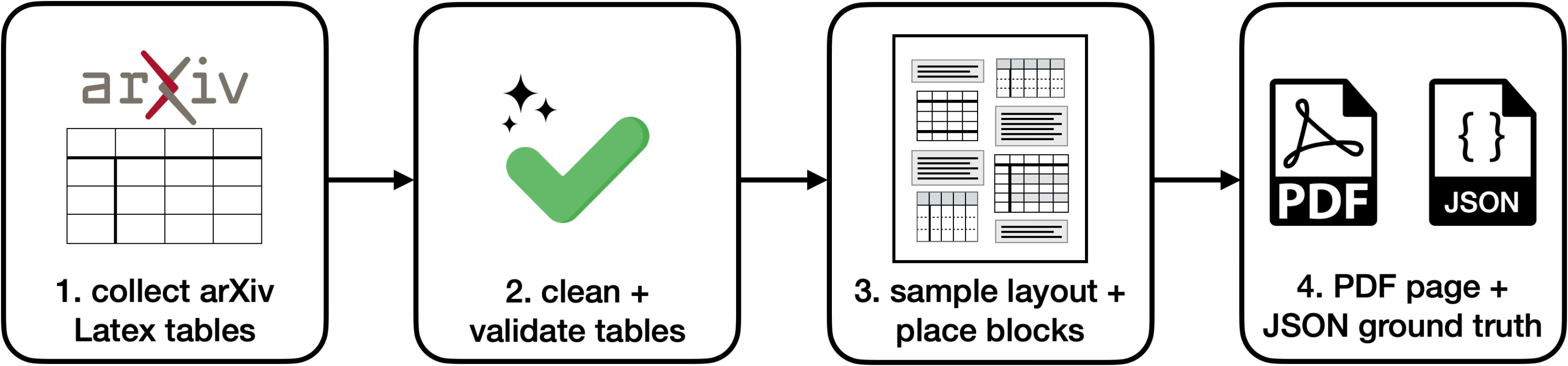}
\caption{Overview of the benchmark generation pipeline: arXiv \LaTeX{} tables are collected, cleaned, and validated, then placed alongside filler text under a sampled layout, yielding a PDF page paired with a JSON ground truth.}
\label{fig:dataset_generation}
\end{figure}

Our corpus addresses the three benchmark concerns from Section~\ref{sec:gaps}: the December~2025 arXiv sample post-dates all 21 evaluated parsers and is regenerable (mitigating contamination), authorial LaTeX sources serve as exact ground truth (avoiding conversion artifacts), and pages contain 4.51~tables on average (table-dense pages).
Figure~\ref{fig:dataset_generation} gives an overview of the pipeline.

Of the 38{,}010 papers published on arXiv in December~2025, we retain the 18{,}676 released under a redistributable Creative Commons license (CC BY, CC BY-SA, CC BY-NC-SA, or CC0) and collect their LaTeX table sources.
All top-level \texttt{tabular}/\texttt{tabular*} environments are extracted, stripped of non-content commands (citations, cross-references, captions), and compiled standalone to verify validity and record rendered dimensions; tables that fail to compile are discarded, yielding 41{,}090 valid tables.
Each table is classified by GPT-5-mini into \emph{simple} (64.0\%, regular grid, no cell merging), \emph{moderate} (24.2\%, limited cell merging), or \emph{complex} (11.7\%, multi-dimensional merging or nested structures).

Each benchmark page is generated by sampling a random layout configuration (document class, font family, page margins, font size, line spacing, and single- or two-column layout) and iteratively appending content blocks, either tables from the extracted pool or variable-length English filler text generated with the Faker library.
Tables are pre-filtered by their recorded dimensions against the remaining page space, scaled to column width via \texttt{adjustbox} when moderately oversized, and placed as non-floating centered blocks to avoid the unpredictable reordering of LaTeX float environments that would complicate ground truth alignment.
The document is recompiled with \texttt{pdflatex} after each addition; blocks that trigger overflow or typesetting warnings are discarded, and the process terminates when no further content fits.
Each page thus yields a PDF alongside the full sequence of placed blocks as ground truth, of which we evaluate the embedded LaTeX tables.

\subsection{Table Matching}
\label{sec:matching}

Before tables can be evaluated, each ground truth table must be matched to its counterpart in the parser output.
This is non-trivial because parsers produce tables in diverse formats (HTML, Markdown, LaTeX, plain text), may split or merge tables, reorder content in multi-column layouts, or fail to recognize tables entirely.

We address this with an LLM-based matching pipeline using Gemini-3-Flash-Preview~\cite{gemini_3}.
Given the ground truth tables (as LaTeX) alongside the full parser output, the model is prompted to return for each ground truth table the corresponding parsed span verbatim, or an empty string if no match is present.
Since LLMs occasionally introduce minor transcription artifacts such as altered whitespace, the returned text is treated as a locator only: a rule-based post-validation step recovers the matched span as a literal substring of the parser output via sliding-window Levenshtein alignment.
The pipeline therefore cannot fabricate matches: every matched span originates verbatim from the parser output.
Manual inspection of the matched spans across all 21 parser outputs confirmed reliable matching behavior across the encountered output formats (HTML, Markdown, LaTeX, plain text).

\section{Metric Meta-Evaluation}
\label{sec:assessment}

Once ground truth and parsed tables have been aligned, the central question becomes how to quantify extraction quality.
This section exposes the limitations of existing rule-based metrics, establishes a human reference, introduces LLM-based semantic evaluation as an alternative, and quantifies how well each approach captures human judgment.

\subsection{Limitations of Rule-based Metrics}
\label{sec:metrics_limitations}

TEDS, GriTS, and SCORE compare structure and content syntactically and cannot separate semantic from merely representational discrepancies.
Figure~\ref{fig:metric_example} illustrates discrepancy patterns we frequently observed in parser outputs.
Several difference types are semantically insignificant yet incur large edit distances:
\begin{itemize}
    \item \hl{clayout!80}{\emph{Structural reorganization}}: parsers flatten multi-level headers or resolve row\-spans, for instance when the output format lacks support for spanning cells (e.g., Markdown, unlike HTML, has no \texttt{colspan}/\texttt{rowspan} mechanism), leading to work\-arounds such as repeating values, concatenating multi-level headers, or inserting empty padding cells.
    \item \hl{cnotation!80}{\emph{Representational equivalence}}: the same content is serialized with different conventions, whether at the symbol level (the Greek letter alpha as the Unicode glyph $\alpha$, the HTML entity \texttt{\&alpha;}, or the LaTeX command \texttt{\textbackslash alpha}) or as placeholders (``\textemdash'' vs.\ ``N/A'')---semantically identical but string-different in every pairing.
    \item \hl{cartifact!80}{\emph{Markup artifact}}: raw markup survives in the cell content, the same bold appearing as \texttt{\textbackslash textbf\{\dots\}}, \texttt{<b>\dots</b>}, or \texttt{**\dots**} across output formats.
\end{itemize}
% fig_metric_example.tex — metric failure case figure
% \input{fig_metric_example} in main document

% Helper: pass-through for raw LaTeX display
\newcommand{\raw}[1]{#1}

\begin{figure}[!htb]
\centering
{\scriptsize\renewcommand{\bfseries}{\fontseries{b}\fontfamily{cmr}\selectfont}%
%
% ============================================================
%  Ground Truth table
% ============================================================
\textbf{(a) Ground Truth}\\[1mm]
\begin{tblr}{
  colspec = {c c X[c] X[c] X[c] X[c]},
  width = \linewidth,
  vlines,
  hline{1,3,5,7} = {},
  hline{2} = {3-6}{},
  hline{4,6} = {2-6}{},
  colsep = 6pt,
  rowsep = 3pt,
  row{1,2} = {bg=clayout},
  cell{1}{1} = {r=2}{font=\bfseries},
  cell{1}{2} = {r=2}{font=\bfseries},
  cell{1}{3} = {c=2}{font=\bfseries},
  cell{1}{5} = {c=2}{font=\bfseries},
  cell{3}{1} = {r=2}{bg=clayout},
  cell{5}{1} = {r=2}{bg=clayout},
  cell{3}{4} = {bg=cnotation},
  cell{3}{5} = {bg=cnotation},
  cell{3}{6} = {bg=cnotation},
  cell{4}{2} = {bg=cnotation},
  cell{4}{3} = {bg=cartifact},
  cell{4}{4} = {bg=cnotation},
  cell{4}{5} = {bg=ccontent},
  cell{5}{4} = {bg=cnotation},
  cell{5}{6} = {bg=cnotation},
  cell{6}{2} = {bg=cnotation},
  cell{6}{4} = {bg=ccontent},
  cell{6}{5} = {bg=cartifact},
}
Group & Method & Task 1 & & Task 2 & \\
 & & Score & Diff & Score & Diff \\
Group 1 & Baseline & 85.0\% & \textemdash & 0.72 $\pm$ 0.03 & \textemdash \\
 & Method $\alpha$ & \textbf{91.2\%} & +6.2 (\textit{p}\,$\leq$\,0.1) & 1.12 & +0.17 \\
Group 2 & Baseline & 79.3\% & \textemdash & 0.65 & \textemdash \\
 & Method $\beta$ & 76.5\% & $-$2.8 & \textbf{1.31} & +0.66 \\
\end{tblr}%
\vspace{3mm}

% ============================================================
%  Parser Output table
% ============================================================
\textbf{(b) Parser Output}\\[1mm]
\begin{tblr}{
  colspec = {c c X[c] X[c] X[c] X[c]},
  width = \linewidth,
  hlines, vlines,
  colsep = 6pt,
  rowsep = 4pt,
  cell{1}{1} = {bg=cheader},
  cell{1}{2} = {bg=cheader},
  cell{1}{3} = {bg=clayout},
  cell{1}{4} = {bg=clayout},
  cell{1}{5} = {bg=clayout},
  cell{1}{6} = {bg=clayout},
  cell{2}{1} = {bg=clayout},
  cell{2}{4} = {bg=cnotation},
  cell{2}{5} = {bg=cnotation},
  cell{2}{6} = {bg=cnotation},
  cell{3}{1} = {bg=clayout},
  cell{3}{2} = {bg=cnotation},
  cell{3}{3} = {bg=cartifact},
  cell{3}{4} = {bg=cnotation},
  cell{3}{5} = {bg=ccontent},
  cell{4}{1} = {bg=clayout},
  cell{4}{4} = {bg=cnotation},
  cell{4}{6} = {bg=cnotation},
  cell{5}{1} = {bg=clayout},
  cell{5}{2} = {bg=cnotation},
  cell{5}{4} = {bg=ccontent},
  cell{5}{5} = {bg=cartifact},
}
Group & Method & \makecell[c]{Task 1\\Score} & \makecell[c]{Task 1\\Diff} & \makecell[c]{Task 2\\Score} & \makecell[c]{Task 2\\Diff} \\
Group 1 & Baseline & 85\% & N/A & \makecell[c]{\raw{\$0.72}\\\raw{\textbackslash pm 0.03\$}} & N/A \\
Group 1 & \makecell[c]{\raw{Method}\\\raw{\$\textbackslash alpha\$}} & \makecell[c]{\raw{\textbackslash textbf}\\\raw{\{91.2\textbackslash\%\}}} & \makecell[c]{\raw{+6.2}\\\raw{(\$p \textbackslash leq 0.1\$)}} & 112 & +0.17 \\
Group 2 & Baseline & 79.3\% & N/A & 0.65 & N/A \\
Group 2 & \makecell[c]{\raw{Method}\\\raw{\$\textbackslash beta\$}} & 76.5\% & 2.8 & \makecell[c]{\raw{\textbackslash textbf}\\\raw{\{1.31\}}} & +0.66 \\
\end{tblr}%
}% end \scriptsize group
\caption{Structural metrics penalize harmless variation while overlooking
critical errors. The parser output~(b) largely preserves the semantics of~(a),
yet incurs heavy edit distance from non-semantic discrepancies
(\hl{clayout!80}{structural reorganization},
\hl{cnotation!80}{representational equivalence},
\hl{cartifact!80}{markup artifact}).
The only meaning-altering errors---a lost decimal and a sign flip
(\hl{ccontent!80}{content error})---barely affect the score.}
\label{fig:metric_example}
\end{figure}

In contrast, the only semantically critical differences, \hl{ccontent!80}{\emph{content errors}} such as a lost decimal point (1.12\,$\to$\,112) and a dropped minus sign ($-$2.8\,$\to$\,2.8), change merely a single character each, contributing minimally to the edit distance.
String-based metrics consequently assign low scores driven by harmless representational variation, while the few-character errors that fundamentally alter the table's meaning are barely reflected.

\subsection{Human Reference Scores}
\label{sec:human_eval}

To establish a human reference for table extraction quality, three independent evaluators each rated the same 518~pairs of ground truth and parsed tables (1,554 ratings in total), sampled across all parsers and table complexities; the mean of the three ratings serves as the reference score for each pair throughout this section.
Each pair was rated on a 0--10 scale reflecting whether the semantic content of the table---all values, headers, and their associations---has been correctly, completely, and unambiguously preserved.
Evaluators received only this high-level criterion rather than a prescriptive scoring rubric, so that any measured agreement with automated metrics reflects shared semantic judgment rather than a shared scoring procedure.
To help annotators catch subtle errors in large tables, the annotation interface displayed each pair in rendered and raw form side by side, with candidate differences pre-identified by an LLM; the final score remained entirely a human decision.

\textbf{Inter-annotator agreement.}
To assess the reliability of the human reference scores, we report agreement among the three annotators.
Krippendorff's~$\alpha$ (interval) is 0.77~\cite{krippendorff2011computing}, with a mean absolute score difference of 1.2 on the 0--10 scale between annotator pairs.
For reference, the leave-one-out (LOO) Pearson correlation of each annotator with the mean of the other two ranges from $r=0.84$ to $0.93$ (average $r=0.89$).

\subsection{LLM-as-a-Judge for Table Evaluation}
\label{sec:llm_judge}

Building on the LLM-as-a-judge paradigm~\cite{zheng2023judging,llm_judge}, we propose using LLMs to assess table extraction quality semantically.
Given a ground truth table and its parsed counterpart, an LLM evaluates the pair on a 0--10 scale against the same high-level criterion given to human evaluators in Section~\ref{sec:human_eval} (semantic preservation of values, headers, and their associations).
The exact prompt is shown in Figure~\ref{fig:judge_prompt} and was iteratively refined on a separate 50-pair set disjoint from the 518 evaluation pairs. It instructs the model to first enumerate concrete error candidates before assigning a final score, which encourages grounded judgments.
We evaluate eight LLMs as judges, spanning proprietary and open-weight families at different price points: Claude Opus~4.6~\cite{claude_opus_4_6}, Gemini~3 Flash Preview, Gemini~3.1 Flash Lite Preview~\cite{gemini_3}, GPT-5.4 nano~\cite{gpt5}, DeepSeek-v3.2~\cite{deepseek_v3_2}, Gemma-4-31b-it and Gemma-4-26b-a4b-it~\cite{gemma_4}, and Mistral-Small-2603~\cite{mistral_small}.

% fig_judge_prompt.tex — LLM-as-a-judge prompt template
% \input{fig_judge_prompt} in main document

\begin{figure}[!htb]
\footnotesize
\setlength{\fboxsep}{6pt}
\noindent\fcolorbox{black!30}{black!3}{%
\begin{minipage}{\dimexpr\linewidth-2\fboxsep-2\fboxrule\relax}
\setlength{\parskip}{3pt}
\setlength{\parindent}{0pt}
You are a strict table evaluator. Your task is to determine if the extracted table correctly represents the ground truth table, focusing on content accuracy, structural preservation, and information completeness. The extracted table was parsed from the rendered table. Disregard LaTeX-specific elements in the ground truth (e.g., comments, styling commands, font formatting) that have no effect on content or structure.

\textbf{Ground Truth Table (LaTeX):} \texttt{\{gt\_table\}}

\textbf{Extracted Table:} \texttt{\{extracted\_table\}}

Evaluate the extracted table using the following criteria:

\hspace*{1em}1.~\textbf{Content accuracy:} Are all cell values, headers, and data correctly preserved?

\hspace*{1em}2.~\textbf{Structure preservation:} Are all rows and columns present, and can each value be unambiguously mapped to its row/column headers? Broken or ambiguous associations count as errors.

\textbf{Note:} Different output formats (markdown, HTML, plain text) are acceptable as long as no information is lost. Apply this key test: Could a reader who sees ONLY the extracted table---without access to the ground truth---unambiguously reconstruct every cell-to-header mapping and all content of the original table? If not, consider the parsing as failed and assign a low score.

First, enumerate up to 5 of the most significant errors and ambiguities found. Then assign a score from 0 to 10, where 10 is a perfect match.
\end{minipage}}
\caption{Prompt template used for the LLM-as-a-judge table evaluation. The placeholders \texttt{\{gt\_table\}} and \texttt{\{extracted\_table\}} are filled with the LaTeX ground truth and the parsed table, respectively. Each judge model is queried with this prompt for every ground-truth/extraction pair.}
\label{fig:judge_prompt}
\end{figure}

\subsection{Correlation with Human Judgment}
\label{sec:correlation}

\begin{figure}[!p]
\centering
\includegraphics[width=\textwidth,height=0.88\textheight,keepaspectratio]{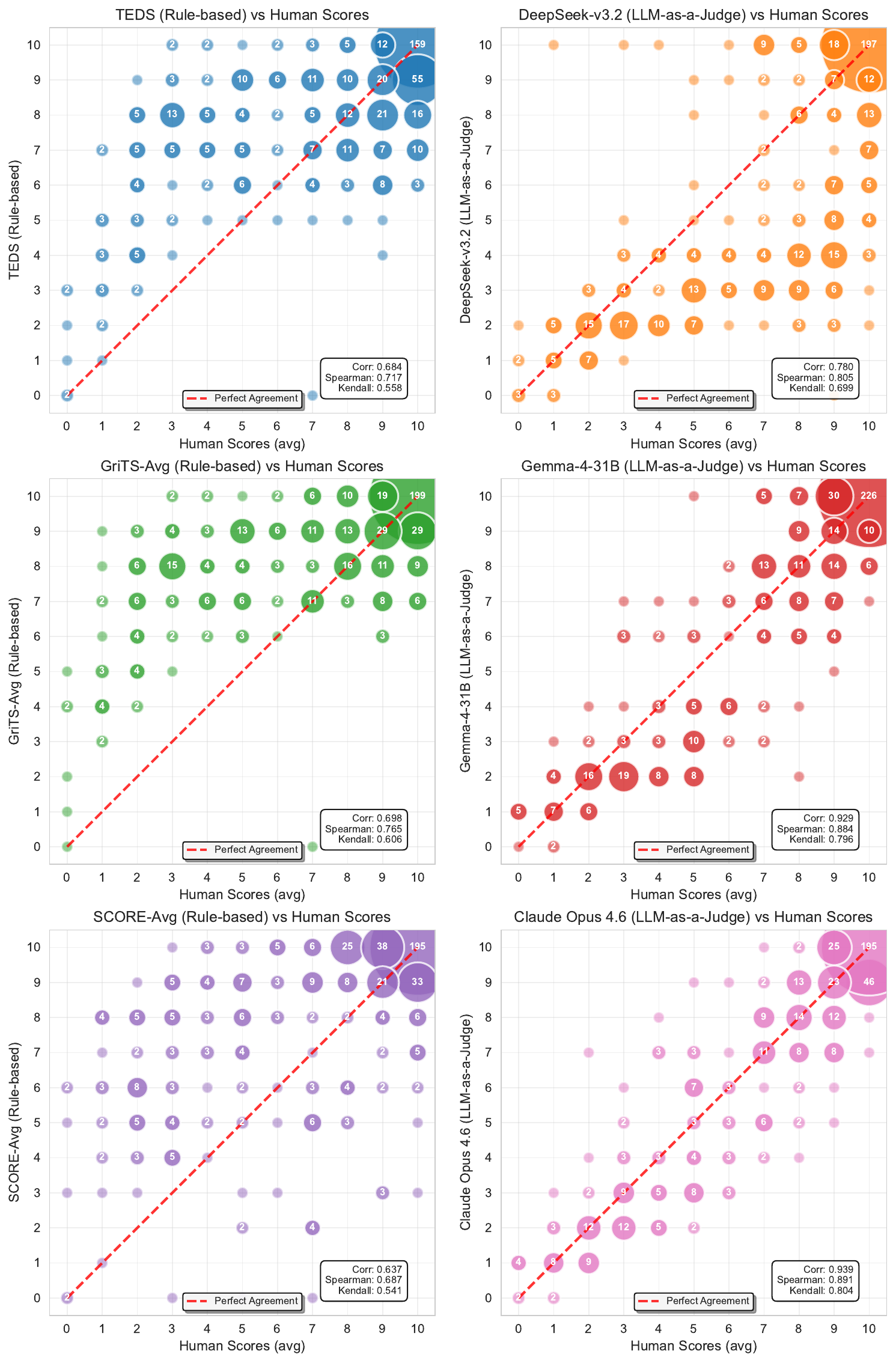}
\caption{Scatter plots comparing automated metrics with human scores. Left column: rule-based metrics (TEDS, GriTS-Avg, SCORE-Avg); right column: LLM judges spanning the score range (DeepSeek-v3.2, Gemma-4-31b-it, Claude Opus~4.6). Bubble size indicates point count.}
\label{fig:correlation}
\end{figure}

We compute Pearson, Spearman, and Kendall correlations between each automated metric and the human reference scores.
All metrics are scaled to a 0--10 range for comparability, and ground-truth tables are converted to HTML where required by the metric (TEDS); Table~\ref{tab:correlation} summarizes the results and Figure~\ref{fig:correlation} visualizes the relationship for a subset of metrics.

\begin{table}[!htb]
\centering
\caption{Correlation of automated metrics with averaged human scores ($n=518$ pairs, three evaluators each). LLM costs: OpenRouter API per 1{,}000 pairs as of April~2026.}
\label{tab:correlation}
\small
\setlength{\tabcolsep}{3pt}
\begin{tabular}{@{}llcccc@{}}
\toprule
\textbf{Metric} & \textbf{Type} & \textbf{Pearson $r$} & \textbf{Spearman $\rho$} & \textbf{Kendall $\tau$} & \textbf{Cost (\$)} \\
\midrule
TEDS & Rule-based & 0.684 & 0.717 & 0.558 & --- \\
$\mathrm{GriTS}_\mathrm{Top}$ & Rule-based & 0.633 & 0.735 & 0.597 & --- \\
$\mathrm{GriTS}_\mathrm{Con}$ & Rule-based & 0.701 & 0.745 & 0.598 & --- \\
GriTS-Avg & Rule-based & 0.698 & 0.765 & 0.606 & --- \\
SCORE Index & Rule-based & 0.558 & 0.684 & 0.561 & --- \\
SCORE Content & Rule-based & 0.642 & 0.657 & 0.524 & --- \\
SCORE-Avg & Rule-based & 0.637 & 0.687 & 0.541 & --- \\
\midrule
Claude Opus~4.6$^\dagger$ & LLM & \textbf{0.939} & 0.891 & \textbf{0.804} & 7.60 \\
Gemma-4-31b-it & LLM & 0.929 & 0.884 & 0.796 & \textbf{0.18} \\
Gemini~3 Flash Preview & LLM & 0.924 & \textbf{0.892} & 0.803 & 0.78 \\
Gemma-4-26b-a4b-it & LLM & 0.909 & 0.861 & 0.766 & 0.54 \\
Gemini~3.1 Flash Lite Preview & LLM & 0.909 & 0.851 & 0.754 & 0.36 \\
GPT-5.4 nano & LLM & 0.809 & 0.799 & 0.683 & 0.28 \\
DeepSeek-v3.2 & LLM & 0.780 & 0.805 & 0.699 & 0.42 \\
Mistral-Small-2603 & LLM & 0.756 & 0.799 & 0.685 & 0.28 \\
\bottomrule
\multicolumn{6}{@{}l}{\footnotesize $^\dagger$\,Also used to generate error hints shown to evaluators; see text.}
\end{tabular}
\vspace{-3mm}
\end{table}

\textbf{Rule-based metrics} achieve only moderate correlation with human judgment.
Since GriTS and SCORE each decompose into separate structure and content sub-metrics, unlike TEDS and LLM-based judges which assess both aspects jointly, we additionally compute their arithmetic means (GriTS-Avg, SCORE-Avg) to enable direct comparison.
All rule-based metrics, whether structure-focused, content-focused, or averaged, fall within a narrow band of $r=0.56$--$0.70$ (Table~\ref{tab:correlation}), confirming the limitations analyzed in Section~\ref{sec:metrics_limitations}.

\textbf{LLM-based evaluation} outperforms all rule-based metrics across every judge: even the weakest (Mistral-Small-2603, $r=0.76$) exceeds the best rule-based metric (GriTS-Con, $r=0.70$), confirming that semantic assessment captures aspects of table quality that string-level comparison misses.
Claude Opus~4.6 leads at $r=0.94$ but may be confounded as it also generated the annotator error hints ($^\dagger$, Table~\ref{tab:correlation}); setting Claude aside, four further judges reach $r\geq0.90$, including the open-weight Gemma-4-31b-it ($r=0.93$).
Any priming effect from Claude generating the hints is bounded: the next two judges not involved in hint generation (Gemma-4-31b-it, Gemini~3 Flash) trail Claude by only $\Delta r \leq 0.02$, well below the $\geq 0.23$ rule-based–vs.–LLM gap.
Bootstrap 95\,\% CIs place the two leading judges (Claude $[0.92,0.95]$, Gemma-4-31b-it $[0.91,0.94]$) within the per-annotator LOO range $[0.91,0.94]$: they reach the upper end of the inter-annotator spectrum without exceeding it, with the remaining gap to the \emph{average} annotator's LOO correlation concentrated on items where annotators themselves disagree.

\textbf{Prompt sensitivity.}
To test whether the LLM advantage hinges on prompt engineering, we re-evaluate the eight judges on the 518~pairs under two variants: a one-line \emph{naive} prompt (``Rate how well the extracted table matches the ground truth table on a scale from 0 to 10.''), and a \emph{tuned (no CoT)} variant identical to Figure~\ref{fig:judge_prompt} but omitting the error-enumeration step.
The advantage stems primarily from semantic understanding rather than prompt engineering (Table~\ref{tab:prompt_sensitivity}): 23 of 24 configurations exceed the best rule-based metric ($\mathrm{GriTS}_\mathrm{Con}=0.701$), with the engineered prompt adding further gains for most judges (median $\Delta r=+0.05$, max $+0.10$).

For the parser benchmark in Section~\ref{sec:experiments}, we adopt Gemma-4-31b-it as it offers correlation on par with Claude Opus~4.6 ($r=0.93$ vs.\ $r=0.94$) at a fraction of its inference cost (\$0.18 vs.\ \$7.60 per 1{,}000 pairs) while running as an open-weight model.

\begin{table}[!t]
\centering
\footnotesize
\setlength{\tabcolsep}{3pt}
\caption{Pearson correlation with averaged human scores across three prompt variants on the same 518 pairs. Parenthetical values show $\Delta r$ relative to the \emph{Tuned} variant.}
\label{tab:prompt_sensitivity}
\begin{tabular}{@{}lccc@{}}
\toprule
\textbf{Model} & \textbf{Tuned} & \textbf{Tuned (no CoT)} & \textbf{Naive} \\
\midrule
Claude Opus~4.6 & 0.939 & 0.933 ($-$0.006) & 0.891 ($-$0.048) \\
Gemma-4-31b-it & 0.929 & 0.916 ($-$0.013) & 0.907 ($-$0.022) \\
Gemini~3 Flash Preview & 0.924 & 0.907 ($-$0.017) & 0.851 ($-$0.073) \\
Gemma-4-26b-a4b-it & 0.909 & 0.777 ($-$0.132) & 0.856 ($-$0.053) \\
Gemini~3.1 Flash Lite Preview & 0.909 & 0.863 ($-$0.046) & 0.807 ($-$0.102) \\
GPT-5.4 nano & 0.809 & 0.795 ($-$0.014) & 0.718 ($-$0.091) \\
DeepSeek-v3.2 & 0.780 & 0.671 ($-$0.109) & 0.779 ($-$0.001) \\
Mistral-Small-2603 & 0.756 & 0.842 ($+$0.086) & 0.803 ($+$0.047) \\
\bottomrule
\end{tabular}
\end{table}

\section{PDF Parser Leaderboard}
\label{sec:experiments}
Using the validated Gemma-4-31b-it judge, we evaluate 21~parsers on 100 synthetic PDF pages containing 451 tables with diverse structural characteristics.

We selected 21 parsers spanning the full spectrum of contemporary document parsing approaches.
Among specialized OCR models, we evaluate Chandra~\cite{chandra}, DeepSeek-OCR~\cite{deepseek_ocr}, dots.ocr~\cite{dots_ocr}, GOT-OCR2.0~\cite{got_ocr2}, LightOnOCR-2-1B~\cite{lightonocr}, Mathpix~\cite{mathpix}, MinerU2.5~\cite{mineru}, Mistral OCR 3~\cite{mistral_ocr}, MonkeyOCR-3B~\cite{monkeyocr}, Nanonets-OCR-s~\cite{nanonets_ocr_s}, and olmOCR-2-7B~\cite{olmocr2025}.
These range from compact end-to-end vision-language models with under 1B parameters (LightOnOCR, DeepSeek-OCR) to full-page decoders built on larger VLM backbones (Chandra on Qwen3-VL, MonkeyOCR) and commercial API services (Mathpix, Mistral OCR 3).

We also evaluate general-purpose multimodal models including Gemini~3 Pro and Flash, Gemini~2.5 Flash~\cite{gemini_3, gemini_2_5}, GLM-4.5V~\cite{glm_4_5v}, Qwen3-VL-235B~\cite{qwen3_vl}, GPT-5 mini and nano~\cite{gpt5}, and Claude Sonnet~4.6~\cite{claude_opus_4_6}; since these models lack a dedicated document parsing mode, they were prompted to convert each page to Markdown with tables rendered as HTML.

Additionally, we include PyMuPDF4LLM~\cite{pymupdf4llm}, a rule-based tool that extracts text directly from the PDF text layer, and the scientific document parser GROBID~\cite{grobid}.

All 100~pages are processed through each parser and the extracted tables are evaluated against ground truth using the LLM-based pipeline described in Sections~\ref{sec:matching} and~\ref{sec:llm_judge}.
Table~\ref{tab:results} reports the resulting scores alongside the approximate cost or time for parsing all 100~pages: API pricing in USD at the time of writing or wall-clock time on a single NVIDIA RTX~4090.
As most models offer multiple deployment options and we did not use a uniform inference framework (e.g., vLLM or Hugging Face Transformers), reported runtimes are rough estimates.
Our code repository provides ready-to-use implementations for all 21~parsers together with the exact prompts, configurations, and software versions used to produce the leaderboard results, enabling full reproducibility.

\begin{table}[!htb]
\centering
\caption{Table extraction performance across 451 tables from 100 synthetic pages, scored 0--10 by Gemma-4-31b-it and broken down by structural complexity, with TEDS scores (0--1 scale) for comparison. Parsers are ranked by overall score.}
\label{tab:results}
\footnotesize
\setlength{\tabcolsep}{4pt}
\begin{tabular}{@{}lccccccc@{}}
\toprule
& \multicolumn{4}{c}{\textbf{LLM Score (0--10)}} & & & \\
\cmidrule(lr){2-5}
\textbf{Parser} & \textbf{Overall} & \textbf{Simple} & \textbf{Moderate} & \textbf{Complex} & \textbf{TEDS} & \textbf{Inference} & \textbf{Cost\,/\,Time} \\
\midrule
Gemini 3 Pro         & \textbf{9.55} & \textbf{9.58} & \textbf{9.57} & 9.49 & 0.85 & API & \$10.00 \\
Gemini 3 Flash       & 9.50 & 9.53 & 9.38 & \textbf{9.61} & 0.85 & API & \$0.57 \\
LightOnOCR-2-1B      & 9.08 & 9.41 & 8.90 & 8.91 & 0.83 & GPU & 30\,min \\
Mistral OCR 3        & 8.89 & 8.92 & 8.69 & 9.07 & \textbf{0.88} & API & \$0.20 \\
dots.ocr             & 8.73 & 9.01 & 8.43 & 8.76 & 0.81 & GPU & 20\,min \\
Mathpix              & 8.53 & 9.32 & 8.40 & 7.77 & 0.74 & API & \$0.35--0.50 \\
Chandra              & 8.43 & 8.96 & 8.14 & 8.15 & 0.77 & GPU & 4\,h \\
Qwen3-VL-235B        & 8.43 & 9.23 & 8.27 & 7.67 & 0.78 & API/GPU & \$0.20 \\
MonkeyOCR-3B         & 8.39 & 8.60 & 8.10 & 8.47 & 0.80 & GPU & 20\,min \\
GLM-4.5V             & 7.98 & 9.19 & 7.59 & 7.00 & 0.78 & API & \$0.60 \\
GPT-5 mini           & 7.14 & 8.03 & 6.82 & 6.48 & 0.68 & API & \$1.00 \\
Claude Sonnet~4.6    & 7.02 & 6.94 & 7.10 & 7.01 & 0.63 & API & \$3.00 \\
Nanonets-OCR-s       & 6.92 & 8.27 & 6.51 & 5.82 & 0.69 & GPU & 50\,min \\
Gemini 2.5 Flash     & 6.85 & 7.93 & 6.52 & 5.94 & 0.72 & API & \$0.40 \\
MinerU2.5            & 6.49 & 7.07 & 6.03 & 6.35 & 0.78 & API/GPU & ---$^\ddagger$ \\
GPT-5 nano           & 6.48 & 7.63 & 6.18 & 5.47 & 0.32 & API & \$0.35 \\
DeepSeek-OCR         & 5.75 & 7.45 & 5.34 & 4.20 & 0.66 & GPU & 4\,min \\
PyMuPDF4LLM          & 5.25 & 6.78 & 4.86 & 3.91 & ---$^\S$ & CPU & 30\,s \\
GOT-OCR2.0           & 5.13 & 5.89 & 4.95 & 4.45 & 0.58 & GPU & 20\,min \\
olmOCR-2-7B          & 4.05 & 4.64 & 3.78 & 3.68 & 0.35 & GPU & 25\,min \\
GROBID               & 2.10 & 2.27 & 1.94 & 2.09 & ---$^\S$ & CPU & 2\,min \\
\bottomrule
\multicolumn{8}{@{}l}{\footnotesize Cost: API pricing (USD) for 100 pages. Time: wall-clock on a single NVIDIA RTX~4090.}\\
\multicolumn{8}{@{}l}{\footnotesize $^\ddagger$\,Tested via free-tier API; also available for local GPU deployment.}\\
\multicolumn{8}{@{}l}{\footnotesize $^\S$\,TEDS not applicable; output lacks tabular structure entirely.}
\end{tabular}
\end{table}

\section{Discussion}

\paragraph{LLM scores vs.\ TEDS.}
The TEDS scores in Table~\ref{tab:results} illustrate the metric limitations discussed in Section~\ref{sec:metrics_limitations} and quantified in Section~\ref{sec:correlation}.
When parsers that frequently fail to detect tables entirely are excluded, TEDS clusters within 22\% of its scale (0.66--0.88), compressing visibly different parsers into a narrow band.
LLM-based scores, by contrast, span 38\% (5.75--9.55), far better reflecting the substantial quality differences visible upon manual inspection.

\paragraph{Parser performance patterns.}
Overall scores range from 2.10 to 9.55, so parser choice can largely determine whether extracted tables are usable.
The top-performing systems are the Gemini~3 models, general-purpose multimodal models rather than dedicated OCR tools, suggesting that broad visual-linguistic capabilities transfer well to table extraction.
Targeted design can still rival much larger models: LightOnOCR-2-1B reaches 9.08 with only 1B~parameters, and dots.ocr (8.73) and MonkeyOCR-3B (8.39) also run on a single consumer GPU, offering competitive self-hosted alternatives to API services.
Rule-based tools (PyMuPDF4LLM, GROBID) require no GPU but lag substantially behind all learning-based approaches.
The complexity breakdown reveals uneven sensitivity to table complexity: drops from simple to complex range from negligible (Gemini~3 Flash even scores higher) to severe (GLM-4.5V: $-$2.19, Qwen3-VL: $-$1.56, Mathpix: $-$1.55), indicating that handling multi-dimensional cell merging remains a key differentiator.
Even the top-scoring Gemini~3 models exhibit errors upon manual inspection (misaligned spanning cells, altered values, incorrect header-cell associations), confirming that accurate table extraction from PDFs remains unsolved.

\begin{figure}[!p]
\centering
\includegraphics[width=\textwidth,height=0.85\textheight,keepaspectratio]{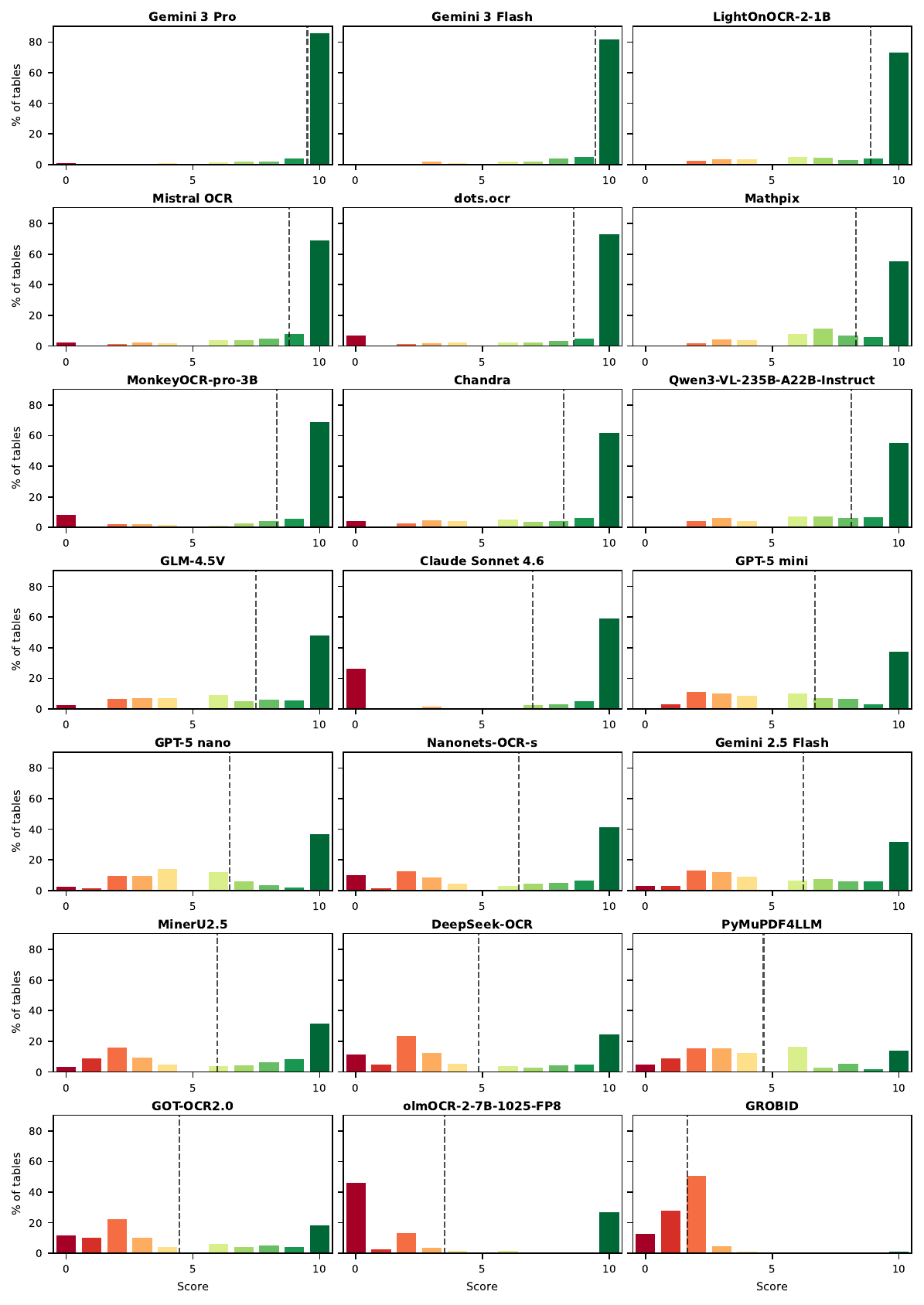}
\caption{Per-parser score distributions across 451 tables. Each subplot shows the percentage of tables receiving each integer score (0--10); the dashed line marks the mean. Parsers are ordered by mean score (top-left to bottom-right).}
\label{fig:score_distributions}
\end{figure}

\paragraph{Score distributions.}
The per-parser histograms in Figure~\ref{fig:score_distributions} expose failure patterns that mean scores obscure.
Top parsers (Gemini~3, LightOnOCR) concentrate $>$70\% of tables at score~10, while Claude Sonnet~4.6 and olmOCR show strongly bimodal distributions: they frequently omit tables entirely (score~0) but extract them near-perfectly when they do.
Mid-tier parsers such as GPT-5~mini and Gemini~2.5~Flash produce broad distributions centered around scores 5--8, indicating pervasive partial errors rather than clean successes or outright failures.
Depending on the application, a missed table may be preferable to a corrupted one, making bimodal parsers with high-quality successes more useful than those with uniformly mediocre output.

\subsection*{Limitations}
Synthetic PDFs do not capture the full diversity of real-world tables (such as scanned documents or non-standard layouts), and the table dataset is sourced exclusively from arXiv, which may bias toward scientific table formats, leaving domains such as financial reports or medical records unrepresented.
While LLM-as-a-judge substantially outperforms rule-based metrics, it is not infallible and introduces additional inference cost compared to rule-based metrics, though this remains modest: scoring all 451~tables costs approximately \$0.10, and a full benchmark run for one parser totals roughly \$1 in API costs.
The leading LLM judges (those at $r\geq0.90$ in Table~\ref{tab:correlation}) correlate more strongly with each other ($r=0.93$--$0.95$) than the human annotators do among themselves ($r=0.81$--$0.91$), indicating a shared LLM prior that should be read as low-noise consensus rather than ground truth.

\subsection*{Future Work}
Future work includes incorporating more diverse document formats and layouts, evaluating parsers' ability to extract information from figures, and extending the benchmark toward holistic document parsing covering tables, formulas, and text jointly.

\paragraph{Code and Data Availability.}
The synthetic PDF generation pipeline, ready-to-use configurations for all 21 parsers, the evaluation pipeline, and the benchmark dataset (100 pages with ground truth) are publicly available.\footnote{\ifanonymous\url{https://anonymous.4open.science/r/pdf-parse-bench-private-7F14}\else\url{https://github.com/phorn1/pdf-parse-bench}\fi}
The meta-evaluation of table extraction metrics, including all metric implementations and the human evaluation study, is provided in a separate repository.\footnote{\ifanonymous\url{https://anonymous.4open.science/r/table-metric-study-5D6F}\else\url{https://github.com/phorn1/table-metric-study}\fi}

\ifanonymous\else
\section*{Acknowledgements}
We thank Sarah Cebulla and Martin Spitznagel for their patience and thoroughness in rating table extraction quality across hundreds of table pairs.
This work has been supported by the German Federal Ministry of Research, Technology, and Space (BMFTR) in the program ``For\-schung an Fach\-hoch\-schu\-len in Ko\-ope\-ra\-tion mit Un\-ter\-neh\-men (FH-Koope\-rativ)'' within the joint project \textit{LLMpraxis} under grant 13FH622KX2.
\fi

\bibliography{references}
\end{document}